# Neural Network Based Epileptic EEG Detection and Classification

*Shivam Gupta[a], Jyoti Meena[b] and O.P Gupta[c]

[a] Department of Computer Science and Engineering , Indian Institute of Information Technology, Sonepat, Haryana, India (Mentor National Institute of Technology, Kurukshetra, Haryana, India )

[b] Department of Computer Engineering , National Institute of Technology, Kurukshetra, Haryana, India

[c] Incharge IT Section, COA, Punjab Agricultural University, Ludhiana, Punjab, India

* Corresponding Author

shivi98g@gmail.com, jtm2067@gmail.com, opgupta@gmail.com

| KEYWORD | ABSTRACT |
|---|---|
|  | *Timely diagnosis is important for saving the life of epileptic patients. In past few years, a lot of treatments are available for epilepsy. These treatments require use of anti-seizure drugs but are not effective in controlling frequency of seizure. There is need of removal of an affected region using surgery. Electroencephalogram (EEG) is a widely used technique for monitoring the brain activity and widely popular for seizure region detection. It is used before surgery for locating affected region. This manual process, using EEG graphs, is time consuming and requires deep expertise. In the present paper, a model has been proposed that preserves the true nature of an EEG signal in form of textual one-dimensional vector. The proposed model achieves a state of art performance for Bonn University dataset giving an average sensitivity, specificity of 81% and 81.4% respectively for classification of EEG data among all five classes. Also for binary classification achieving 99.9%, 99.5% score value for specificity and sensitivity instead of 2D models used by other researchers. Thus, developed system will significantly help neurosurgeons in the increase of their performance.* |

## 1. Introduction

Epilepsy is a disease that affects mostly in initial phase of life. It is neurological disease common worldwide. It is a long lasting disease. This disease cannot spread by direct physical contact. The main distinctive trait of epilepsy is seizure. Seizure are small periods of reflexive shaking of body parts like legs and arms (Abbasi et al., 2019). With recent development in the field of technology many new treatments are available. These treatment involve use of anti-seizure medicine tablets. But these are not effective in controlling frequency of seizure (Abedin et al., Sept 2019). So only available treatment left is to remove affected part of brain using surgery. Electroencephalogram (EEG) are used for analyzing electrical brain activity (Mao et al., Jan 2020). EEG consist of electrodes that take signal from brain. These waves are useful in locating region affected by epilepsy in brain. In maximum cases manual examination of low quality EEG graphs is done (Acharya et al., 2018). This requires a lot of pre expertise. There is no standardized way of recording EEG across countries. This reduces advice from international experts (Bhagat et al., 2019). This all thus increases pressure on neurosurgeons and make surgery cumbersome. So there is need of automation using Artificial Intelligence for classifying epileptic data. This will help easy classification of EEG signal into classes of medical significance. In the present study, a model has been proposed that achieves a state of art performance for Bonn University dataset. It is giving an average sensitivity, specificity of 81% and 81.4% respectively for classification of EEG data into five classes. Also for binary classification, model has achieved 99.9%, 99.5% score value for specificity and sensitivity which is better than the other researchers. Further the true nature of the raw textual data is also lost while development of two dimensional (2D) models used by them. Thus, it will assist neurosurgeons and will help in easy location of affected part of brain by development of internet of things (IoT) based prediction and diagnosis devices

## 2. Literature Review

Lian et al. reported average sensitivity, specificity of 99.5% and 99.6% respectively when convolutional neural network (CNN) was fed with pair of EEG signals to form a 2-dimensional (2D) matrix (Lian et al., Feb 2020). In 2019, Choi et al. came up with multi-scale 3D CNN with deep neural network (DNN) model which achieved a sensitivity and specificity of 97% and 99.3% respectively (Choi et al., 2019). Huang et al. proposed an attention based CNN-BiRNN (bidirectional recurrent neural network) to classify between healthy and unhealthy patients. It achieved an average sensitivity and specificity of 93.94%, 92.88% respectively (Huang et al., 2019). Wei et al. came up with a 12 layered CNN in 2019 that classified raw data with sensitivity and specificity of 70.68%, 92.36% for all five classes (Wei et al., 2019). Liu et al. proposed a 2D CNN that gave an accuracy of 64.5% for five classes (Liu et al., Jan 2020). The proposed CNN was fed with multi bio-signals along with EEG signals. Thanaraj et al. reported accuracy of 92% using CNN for binary classification. The CNN was fed with EEG signals transformed into RGB images using Gramian Angular Summation Field (GASF). The developed CNN consisted of pre-trained standard architectures namely AlexNet, VGG16 and VGG19 (Thanaraj et al., March 2020).

From literature review, it is found that most of work is carried out by converting raw textual data into 2D form. It is either done by pairing EEG vectors or by converting textual data into images and spectrographs. This conversion into 2D form destroys rue nature of raw textual data. This results in performance degradation. Also most of work is carried out around binary classification in which feature extraction is easier as compared to classification for all five classes.

## 2.1. Dataset

The Epileptic Seizure Recognition Dataset is taken from UCI Machine learning repository. The dataset is prepared by Department of Epileptology, Bonn University (Andrzejak et al., 2001). It is free and open to use for research work. It contains raw textual EEG signal data. Data is recorded for 23.6 seconds (Yeola et al., 2019). It includes 11500 samples of length 178 and 1 column for label class. The description of all classes are presented in tabular form in Table 1.

*Table 1.  Description of dataset classes*

| Class Name | No of Samples | Medical Class Label | Description |
|---|---|---|---|
| Z | 2300 | healthy (open) | Healthy person having open eyes |
| O | 2300 | healthy (close) | Healthy person having close eyes |
| N | 2300 | inter-ictal | The signals between two consecutive ictal |
| D | 2300 | pre-ictal | The signals before on-site of seizure |
| S | 2300 | ictal | Signals during seizure |

# 3. Research Methodologies

In the present study, complete epilepsy diagnosis process is proposed to divide into two jobs for progressive improvement in the training of the model. It is based on the dataset classifications shown in the Table 1 that are significant in locating and treating of the cortical region of brain affected by Epilepsy.

## 3.1. Training Methodology

The model is trained for two different jobs, which are as follows:-

Job 1:  To classify between healthy person (AB) and person during seizure (E).

Job 2:  To classify between all the five states i.e. healthy (open), healthy (closed), inter-ictal, pre-ictal and ictal.

## 3.2. Proposed Model

The proposed model is implemented using pytorch library in python. The 76% of total dataset of 11500, is used for training purpose. The validation and testing is using 12%, 12% respectively of left out dataset. A total of 20 epoch are used for training the model. An epoch refers to complete traversal of whole dataset. The model having least loss is saved for future testing.  The loss of a model helps in evaluating how well a model perform after each iteration of optimization.  The loss function used in evaluating model is Cross Entropy. The model mainly consist of two blocks viz. basic block and proposed model block. The dataflow diagram of both blocks are shown in Figure 1 and Figure 2. The skip connections has been used in basic block to overcome the vanishing gradient as shown in the Figure 1.

In Binary classification i.e. job1 the cross entropy is calculated using formula:

$$Loss = -(y \log(p) + (1 - y) \log(1 - p)) \qquad (1)$$

In Multi class classification i.e. job2 the loss is calculated for each class and results are summed up:

$$Loss = -\sum_{c=1}^{M} y_c \log(p_c) \qquad (2)$$

$M$ : Number of classes
log: The natural log
p: predicted probability of class c
y: binary indicator (0 or 1) if label c is correct observation

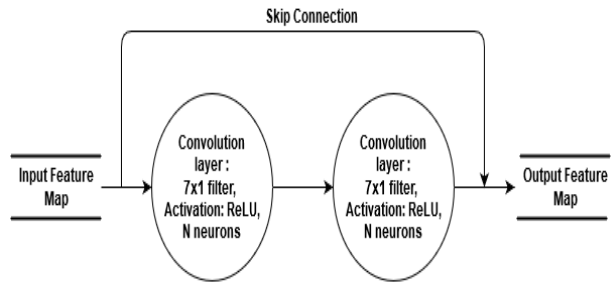

*Figure 1. Dataflow diagram of basic block.*

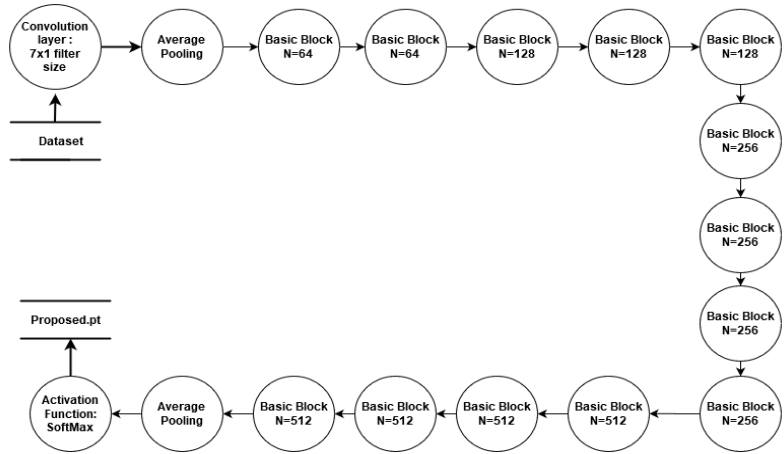

*Figure 2. Dataflow diagram of proposed model block.*

In the Figure 2, Number of Neurons are represented with N and various sequential basic blocks as shown in the Figure 1, are used with different number of neurons.

## 3.3. Algorithms

The pictorial representation of the algorithms for both basic block and proposed model is shown in Figure 3 and Figure 4 respectively. The algorithm for proposed model helps in feature extraction. It also helps in predicting the label for input EEG vector. It calls basic block algorithm within to complete the process.

**I.      Algorithm for Basic Block with Input Vector V and Number of Neurons N.**

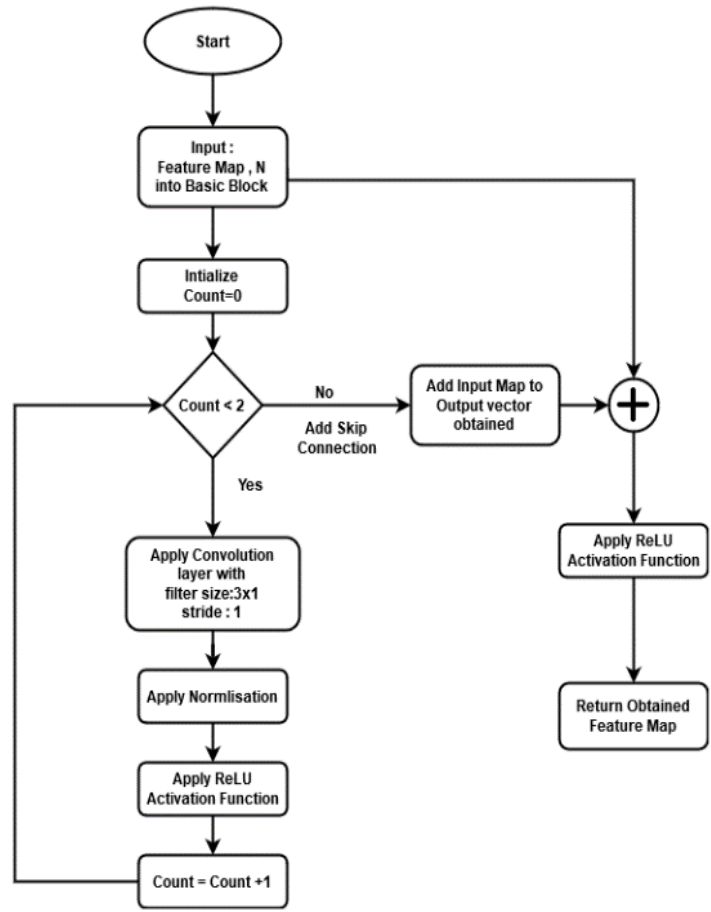

*Figure 3.  Algorithm for basic block.*

**II.    Algorithm for proposed model with Input EEG Signal vector E, and Number    of Classes C**

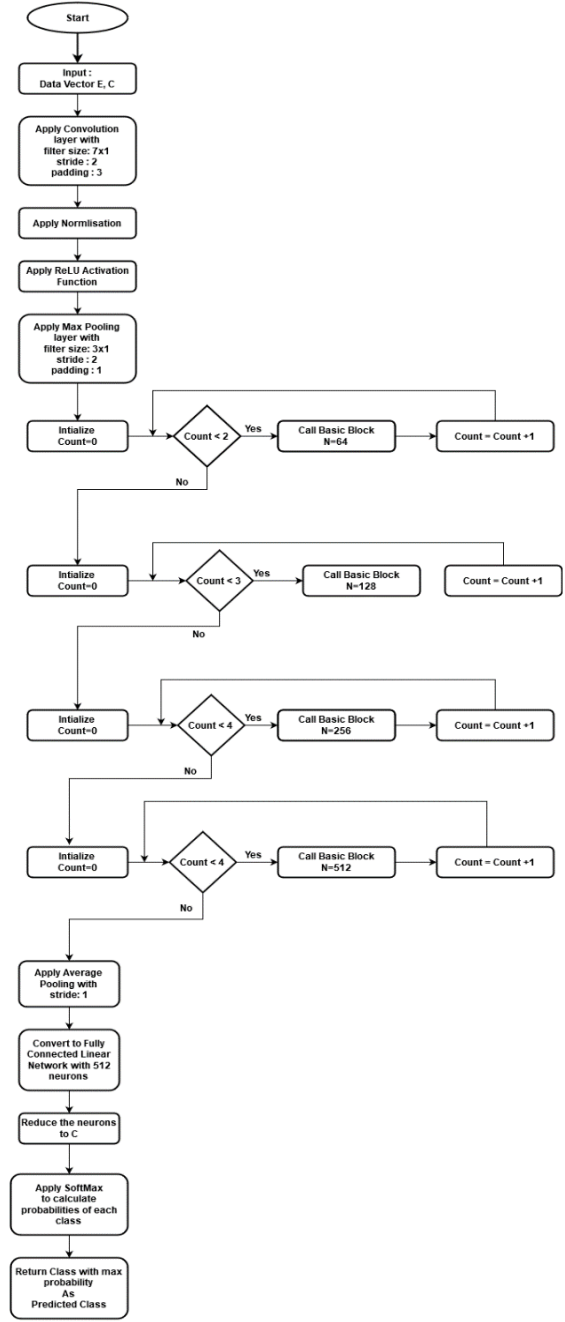

*Figure 4.  Algorithm for proposed model.*

# 4. Results and Discussion

In the proposed model is executed for 20 epochs and best model is saved in external file for future testing and diagnosis. The metrics other than loss function 'Cross Entropy' used for evaluating best fit model are listed below:

- Specificity (Precision): helps in evaluating models ability to predict true negative of each class. It has been evaluated on proposed model using formula:

$$Specificity \ (For \ class \ c) = \frac{\sum TP_c}{\sum FP_c + \sum TP_c} \qquad (3)$$

- Sensitivity (Recall): helps in evaluating models ability to predict true positive of each class. It has been evaluated on proposed model using formula :

$$Sensitivity \ (For \ class \ c) = \frac{\sum TP_c}{\sum FN_c + \sum TP_c} \qquad (4)$$

- F1 score: helps in conveying the balance between precision and recall. It is calculated using harmonic between two by formula shown below:

$$F1 \ score \ = \frac{2 * specificity * senstivity}{specificity + senstivity} \qquad (5)$$

## 4.1. Proposed Model Execution

**Job 1:** To classify between healthy person (AB) and person during seizure (E).

During the training session of the proposed model for the Job 1, the best fit model is found during 9th epoch. The trend of training and validation loss of proposed model when executed for job 1 is shown in Figure 5 (a). Also the classification report for all classes in job 1 is shown in Figure 5 (b). The report contains the value of specificity (precision), sensitivity (recall) and F1 score for each class. The average sensitivity, specificity score for job 1 is 0.99 and 0.995 respectively. The training loss and validation loss for best fit model is shown in tabular form in Table 2.

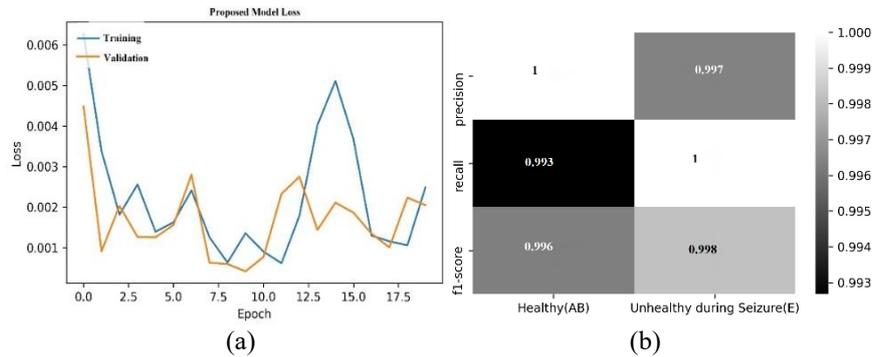

(a)                                    (b)

*Figure 5. (a) Training vs. Validation loss for job 1   (b) Classification report for job 1.*

**Job 2:** To classify between all the five states i.e. healthy (open), healthy (closed), inter-ictal, pre-ictal and ictal.

During the training session of the proposed model for the Job 2, the best fit model is found during 14th epoch. The trend of training and validation loss of proposed model when executed for job 2 is shown in Figure 6 (a). Also the classification report for all classes in job 2 is shown in Figure 6 (b). The report contains the value of specificity (precision), sensitivity (recall) and F1 score for each class. The average sensitivity, specificity score for job 1 is 0.814 and 0.81 respectively. The training loss and validation loss for best fit model is shown in tabular form in Table 2.

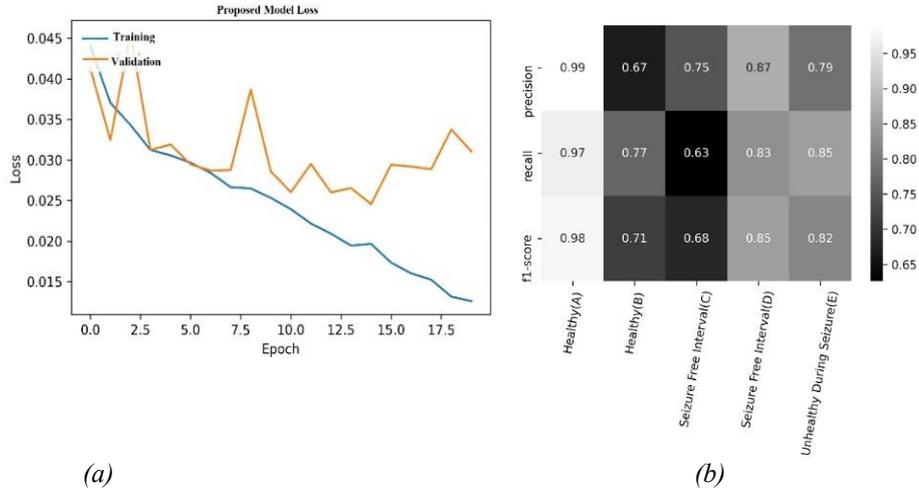

*(a)*                      *(b)*

*Figure 6. (a) Training vs. Validation loss for job 2     (b) Classification report for job 2.*

## 4.2. Comparative analysis with Existing Architectures

Table 2 compares the Training loss and Validation testing loss of proposed model with existing standard models. The existing standards models were also executed on the same dataset to have normalized loss for all the jobs viz. job 1and job 2. It can be observed that the proposed model is working better (having least loss value) than standard models hyper tuned to work on the same dataset.

*Table 2. Training and Validation Testing loss comparison for job 1 and job 2*

| Architecture | Job 1 | | Job 2 | |
|---|---|---|---|---|
| | Training Loss(x$10^{-3}$) | Validation Loss(x$10^{-3}$) | Training Loss(x$10^{-3}$) | Validation Loss(x$10^{-3}$) |
| LeNet | 1.2524 | 2.3996 | 59.4740 | 57.5233 |
| AlexNet | 0.5046 | 0.6056 | 28.5683 | 26.3285 |
| VGG13 | 0.4309 | 0.4324 | 25.0354 | 25.7851 |
| DenseNet 121 | 0.4310 | 0.4330 | 25.4458 | 24.4857 |
| **Proposed Model** | **0.4186** | **0.6310** | **24.5594** | **23.6738** |

## 4.3. Comparative analysis with Existing Methods

Table 3 compares the average specificity and sensitivity over all classes of each job. It can be observed that the proposed model is having better precision (specificity) and sensitivity (recall) over existing methods. The proposed model has better balance between specificity and sensitivity in comparison to Wei et.al. in job2. Hence proposed model achieves a state of art performance.

*Table 3. Average specificity and sensitivity for job 1 and job 2*

| Methods | Job 1 | | Job 2 | |
| --- | --- | --- | --- | --- |
| | Average Specificity | Average Sensitivity | Average Specificity | Average Sensitivity |
| Wei et al. | - | - | 92.36% | 70.68% |
| Huang et al. | 93.94% | 92.88% | - | - |
| Choi et al. | 99.3% | 97% | - | - |
| Lian et al. | 99.0% | 99.0% | - | - |
| **Proposed Model** | **99.9%** | **99.5%** | **81.4%** | **81%** |

# 5. Conclusion

In the proposed model, one dimensional Convolutional Neural Network (CNN) is used which overcomes the problem of vanishing gradient by using skip connections in basic blocks and also protects the true nature of input data (textual). In comparison to existing work, that are converting the data into two dimensional either in the form of matrix or converting the data into images and graphs.

It is concluded from the Tables 2 that the proposed model is better than the existing standard models (approx. 50% less loss). With a balanced F1 score, the proposed model has a good amount of balance between sensitivity and specificity. Due to difficult feature extraction in classification of EEG data into all five classes, most of work is carried for binary classification. But the proposed model performs better for classification into all five classes along with the binary classification. From Table 3 it is clear that proposed model achieves a state of art performance for dataset, taken from Bonn University, giving an average sensitivity, specificity of 81% and 81.4% respectively for classification among all five classes. Further, binary classification is achieving 99.9%, 99.5% score value for specificity and sensitivity. Thus the present study is important which classifies EEG data into all five classes. These all five classes have great significance in the field of medical science for further development of internet of things (IoT) based prediction and diagnosis devices.

# References

Abbasi, M. U., Rashad, A., Basalamah, A., & Tariq, M. (2019). Detection of Epilepsy Seizures in Neo-Natal EEG Using LSTM Architecture. *IEEE Access*, *7*, 179074-179085.

Abedin, M. Z., Akther, S., & Hossain, M. S. (2019, September). An Artificial Neural Network Model for Epilepsy Seizure Detection. In 2019 5th International Conference on Advances in Electrical Engineering (ICAEE) (pp. 860-865). IEEE.

Acharya, U. R., Oh, S. L., Hagiwara, Y., Tan, J. H., & Adeli, H. (2018). Deep convolutional neural network for the automated detection and diagnosis of seizure using EEG signals. Computers in biology and medicine, 100, 270-278.

Andrzejak, R. G., Lehnertz, K., Mormann, F., Rieke, C., David, P., & Elger, C. E. (2001). Indications of nonlinear deterministic and finite-dimensional structures in time series of brain electrical activity: Dependence on recording region and brain state. *Physical Review E*, *64*(6), 061907.

Bhagat, P. N., Ramesh, K. S., & Patil, S. T. (2019). An automatic diagnosis of epileptic seizure based on optimization using Electroencephalography Signals. *Journal of Critical Reviews*, *6*(5), 200-212.

Choi, G., Park, C., Kim, J., Cho, K., Kim, T. J., Bae, H., ... & Chong, J. (2019, January). A novel multi-scale 3D CNN with deep neural network for epileptic seizure detection. In *2019 IEEE International Conference on Consumer Electronics (ICCE)* (pp. 1-2). IEEE.

Huang, C., Chen, W., & Cao, G. (2019, November). Automatic Epileptic Seizure Detection via Attention-Based CNN-BiRNN. In *2019 IEEE International Conference on Bioinformatics and Biomedicine (BIBM)* (pp. 660-663). IEEE.

Lian, J., Zhang, Y., Luo, R., Han, G., Jia, W., & Li, C. (2020). Pair-Wise Matching of EEG Signals for Epileptic Identification via Convolutional Neural Network. *IEEE Access*, *8*, 40008-40017.

Liu, Y., Sivathamboo, S., Goodin, P., Bonnington, P., Kwan, P., Kuhlmann, L., ... & Ge, Z. (2020, February). Epileptic Seizure Detection Using Convolutional Neural Network: A Multi-Biosignal study. In *Proceedings of the Australasian Computer Science Week Multiconference* (pp. 1-8).

Mao, W. L., Fathurrahman, H. I. K., Lee, Y., & Chang, T. W. (2020, January). EEG dataset classification using CNN method. In *Journal of Physics: Conference Series* (Vol. 1456, No. 1, p. 012017). IOP Publishing.

Thanaraj, K. P., Parvathavarthini, B., Tanik, U. J., Rajinikanth, V., Kadry, S., & Kamalanand, K. (2020). Implementation of Deep Neural Networks to Classify EEG Signals using Gramian Angular Summation Field for Epilepsy Diagnosis. *arXiv preprint arXiv:2003.04534*.

Wei, Z., Zou, J., Zhang, J., & Xu, J. (2019). Automatic epileptic EEG detection using convolutional neural network with improvements in time-domain. *Biomedical Signal Processing and Control*, *53*, 101551.

Yeola, L. A., & Satone, M. P. (2019). Deep Neural Network for the Automated Detection and Diagnosis of Seizure using EEG Signals.

http://archive.ics.uci.edu/ml/datasets/Epileptic+Seizure+Recognition downloaded, Jan 2020.